\documentclass[a4paper, 10pt, conference]{ieeeconf}
\IEEEoverridecommandlockouts
\usepackage[nolist]{acronym}
\usepackage{tabularx}
\usepackage{colortbl}
\usepackage{tcolorbox}
\usepackage{graphics} % for pdf, bitmapped graphics files
\usepackage{graphicx}
\usepackage{url}
\usepackage{float}
\usepackage{subfigure}
\usepackage{lettrine}

\usepackage{algorithm}
\usepackage{algorithmic}
\usepackage[normalem]{ulem}
\usepackage{soul}

\usepackage{pgfgantt}
\usepackage{tikz}
\usepackage{pgfgantt}
\usepackage{pdflscape}

\usepackage{textcomp}
\usetikzlibrary{arrows,backgrounds,calc,patterns,shapes,positioning,shadows,trees,babel}
\usepackage{hhline}
\usepackage{graphicx}

\usepackage{amssymb}% http://ctan.org/pkg/amssymb
\usepackage{pifont}% http://ctan.org/pkg/pifont
\newcommand{\cmark}{\ding{51}}%
\newcommand{\xmark}{\ding{55}}%

\usepackage{fancyhdr}
\usepackage{amsmath}

\makeatletter
\newcommand\fs@ruled@notop{\def\@fs@cfont{\bfseries}\let\@fs@capt\floatc@ruled

  \def\@fs@pre{}%
  \def\@fs@post{\kern2pt\hrule\relax}%
  \def\@fs@mid{\kern2pt\hrule\kern2pt}%
  \let\@fs@iftopcapt\iftrue}
\renewcommand\fst@algorithm{\fs@ruled@notop}
\makeatother
\restylefloat{algorithm}

\usepackage{amsmath} % assumes amsmath package installed
\usepackage{amssymb}  % assumes amsmath package installed
\usepackage{color, soul}

\makeatletter
\let\NAT@parse\undefined
\makeatother
\usepackage{hyperref}
\hypersetup{
  bookmarks=true,%                   %%% generate bookmarks ...
  bookmarksnumbered=true,%           %%% ... with numbers
  bookmarksopen=true,%               %%% show complete tree of bookmarks
  hypertexnames=true,%              %%% needed for correct links to figures !!!
  breaklinks=true,%                  %%% break links if exceeding a single line
  colorlinks=true,
  linkcolor=black,
  citecolor=green,
  pagebackref=true,
  urlcolor=blue}

\title{\LARGE \bf A Hybrid SLAM and Object Recognition System for Pepper Robot}
\author{Paola Ard\'on$^{*}$, Kaisar Kushibar$^{*}$, Songyou Peng$^{*}$
        \thanks{Paola Ard\'on is with Edinburgh Robotics Center, UK.}
        \thanks{Kaisar Kushibar is with VICOROB, UdG, Spain.}
        \thanks{ Songyou Peng is with I2R, A*STAR, Singapore.}
        \thanks{* \textbf{All authors contributed equally to this work}. The project was done while they were students in Computer Vision and Robotics master (VIBOT) at Heriot-Watt University.}        
}

\begin{document}
\begin{acronym}[ransac]
  \acro{SLAM}{Simultaneous Localisation and Mapping}
  \acro{ERL}{European Robotics League}
  \acro{SIFT}{Scale Invariant Feature Transform}
  \acro{RANSAC}{Random Sample Concensus}
  \acro{SURF}{Speed Up Robust Features}
  \acro{ORB}{Oriented FAST Rotated BRIEF}
  \acro{BRISK}{Binary Robust Invariant Scalable Keypoints}
\end{acronym}

\maketitle
\thispagestyle{plain}
\pagestyle{plain}

\begin{abstract}
Humanoid robots are playing increasingly important roles in real-life tasks especially when it comes to indoor applications. 
Providing robust solutions for the tasks such as indoor environment mapping, self-localisation and object recognition are essential to make the robots to be more autonomous, hence, more human-like. The well-known Aldebaran service robot Pepper is a suitable candidate for achieving these goals. In this paper, a hybrid system combining Simultaneous Localisation and Mapping (SLAM) algorithm with object recognition is developed and tested with Pepper robot in real-world conditions for the first time. The ORB SLAM 2 algorithm was taken as a seminal work in our research. Then, an object recognition technique based on Scale-Invariant Feature Transform (SIFT) and Random Sample Consensus (RANSAC) was combined with SLAM to recognise and localise objects in the mapped indoor environment. The results of our experiments showed the system's applicability for the Pepper robot in real-world scenarios. Moreover, we made our source code available for the community at \url{https://github.com/PaolaArdon/Salt-Pepper}.

\end{abstract}
 
% The key words
\begin{keywords}
Pepper robot, Visual SLAM, Object recognition, ROS.
\end{keywords}

%%%%%%%%%%%%%%%%%%%%%%%%%%%%%%%%%%%%%%%%%%%%%%%%%%%%%%%%%%%%%%%%%%%%%%%%%%%%%%%%
% Sections
\section{Introduction \label{sec:intro}}

The technological advancements in the past decades significantly improved the quality of our daily life.
On an attempt to making humans lives more comfortable and independent, a significant step has been made in the robotics and computer vision field -- the development of humanoid robots.

In comparison with the rest of humanoid robots in the market, Pepper, developed by Aldebaran and Softbank, is affordable and has an open platform that allows developers to enhance its capabilities and implement applications to make the robot useful for everyday tasks~\cite{brown2013meet}.

\subsection{Problem statement and objectives}
    The development of humanoid robots is still a relatively new technological field. Therefore, research is still being done on the subject.
    In order to achieve a higher level of interest and fresh ideas in the area, many competitions are organised around the world such as the \ac{ERL}~\cite{erl}. One of the purposes of this competition is to develop technological applications that will help elderly people to live longer independently at home~\cite{pepper243documentation}. 
    Pepper comes with many built-in functions, some of them being \textit{learn home, object recognition} and \textit{go to goal}. Even though these functions prove to be useful, they are limited in various ways.
    For example, the built-in function \textit{learn-home} requires the environment to be less than $2m^2$. 
    In order to cope with the rules of the \ac{ERL} Service Robots competition and overcome the limitations of Pepper, we set the primary objective: to develop a hybrid system for Pepper that integrates the object recognition into \ac{SLAM}.
    Another limitation that makes the project more challenging is poor sensors that come with the Pepper robot. Since both visual \ac{SLAM} and object recognition use optical cameras, we are giving some characteristics of the used sensors~\cite{pepper243documentation}:
    \begin{itemize}
    \item \textit{RGB camera:} located on the forehead and has a resolution of $320\times 240$ at $5$ frames per second (fps). 
    \item \textit{Depth camera:} located behind Pepper's eyes with a resolution of $320\times240$ at $5$ fps.
    \end{itemize}

\subsection{Contributions and outline}

    Currently, many applications have been developed on the robot family of Aldebaran and Softbank. However, to the best knowledge of the authors, visual SLAM and its combination with a robust object recognition algorithm have never been done on Pepper robot before.
    Our main contributions are the following: 
    \begin{enumerate}
        \item For the first time, a visual \ac{SLAM} algorithm is successfully applied in the Pepper robot, so it is no longer just limited to a small environment. 
        \item We present an accurate and robust object recognition algorithm for Pepper.
        \item We build a hybrid system which integrates the object recognition and \ac{SLAM} into a unified framework. The recognised objects are marked in the map, and the map can be saved and reused. 
        \item The framework has been tested with the Pepper robot in real-world conditions. The demo video is available at \url{https://youtu.be/evFsnWH_bpY}.
        \item We also make our implementation code available for the community at \url{https://github.com/PaolaArdon/Salt-Pepper}.
    \end{enumerate}
    The paper's structure is as follows: Section~\ref{sec:relatedwork} discusses many modern object recognition and \ac{SLAM} methods. The theory of our object recognition method and \ac{SLAM} with Pepper robot are shown in sections~\ref{sec:recognition} and \ref{sec:slam} respectively. Then, the integration of the two functionalities into one system is described in section~\ref{sec:controlAndArch}, which is followed by the results in section~\ref{sec:results} and final remarks as well as future works in section~\ref{sec:final-remarks}.
\section{Related Work} \label{sec:relatedwork}

Before going into details about the used algorithms for the implementation, it is useful to review some of the general concepts in the field. Object recognition and SLAM have been active research fields over the last decades. Some of the related works are reviewed in this section.

\subsection{Object recognition}
    
    Object recognition relates to the problem of identifying an object in an image. In general, the algorithms can be divided into two main streams: appearance-based methods and feature-based methods.
    
    %%%%%%% Appear-based recognition
    An \textbf{appearance-based} recognition method is based on directly using example images (or templates) to perform recognition tasks. 
    Sung \emph{et~al.} \cite{sung1998example} introduced a method where the edges in both, the current frame and template images from the database, are extracted. Then, different sliding windows with various scales are employed to find the object with the highest similarity measures.
    Swain and Ballard in~\cite{swain1991color} initially showed how object recognition could be performed by comparing the colour histogram. 
    Schiele and Crowley~\cite{schiele1996object} applied histograms of receptive fields proposed by Koenderink and van Doorn~\cite{koenderink1992generic} and the recognition result is enhanced with the usage of the Gaussian derivative or the Laplacian operator at multiple scales. 
    Linde and Lindeberg~\cite{linde2004object} generalised the idea of the receptive field histogram to a higher dimensionality.
    Histograms of wavelet coefficients are proved to be a useful tool for the recognition of cars and faces in~\cite{schneiderman2000statistical}. 
    
    Appearance-based methods are usually robust to a particular type of object characteristics, depending on which information is extracted for the comparison between the templates and the objective image. 
    However, they are usually computationally expensive and sensitive to many variations. 
    In contrast, feature-based recognition methods offer a solution to the mentioned problems.
    
    %%%%%%- Featured-based recognition
    The objective of \textbf{feature-based} recognition algorithms is to find feasible matches between the features extracted from the database images and from the target image. Some of the commonly used features in object recognition are: Shape Context~\cite{belongie2002shape}, Haar-Like feature~\cite{viola2001rapid}, \ac{SIFT}~\cite{lowe1999object}, \ac{SURF}~\cite{Bay06surf:speeded}, \ac{ORB}~\cite{rublee2011orb}, \ac{BRISK}~\cite{leutenegger2011brisk}.
    
    %The comparisons of the recognition performance using various features will be shown in detail in the results and discussion part (Section~\ref{sec:results}) of this document.
    In recent years, Artificial Neural Networks -- in particular, Convolutional Neural Networks (CNN) showed promising results for object detection and recognition problems. Unlike the traditional methods that employ hand-crafted features as described above, the CNNs learn the features from the observed data \cite{lecun1998gradient}.
    % In the deep learning era, many robust deep frameworks have also been introduced 
    Recent studies showed that deep CNN architectures are able to outperform the classical algorithms with higher accuracy for object recognition \cite{girshick2015fast,liu2016ssd,redmon2016you}. However, we did not consider using such approaches in this study, as the CNN-based methods rely on the computational power of GPUs which are not currently available for the Pepper robot.
    % \textcolor{red}{\textbf{ADD SOME CNN STATE OF THE ART FOR OBJECT RECOGNITION HERE AND SAY: ALTHOUGH THE DEEP LEARNING METHODS PROVIDE BETTER ACCURACY THAN THE TRADITIONAL METHODS SHOWN ABOVE, WE DID NOT CONSIDER USING SUCH METHODS WITH THE PEPPER ROBOT AS CNN BASED APPROACHES RELY ON GPUs, WHICH IS NOT CURRENTLY AVAILABLE FOR THE PEPPER ROBOT.}}

\subsection{Simultaneous localisation and mapping (SLAM)}
    In this section, we are going to review some of the state-of-the-art algorithms for \ac{SLAM} that we have considered in our project.

    \subsubsection*{\textbf{Extended Kalman filter \ac{SLAM} (EKF-SLAM)}} 
        In EKF-\ac{SLAM} the map is represented with large vector stacking sensors and landmarks states which is modelled by a Gaussian variable~\cite{montemerlo2002fastslam}. Maximum likelihood algorithm is used for data association.
   
        Some of the advantages of EKF-\ac{SLAM} is that it is relatively easy to implement and is efficient when working with a small number of features and distinct landmarks. On the other hand, the complexity is quadratic with respect to the number of features, it does not guarantee convergence in non-linear and/or non-gaussian cases, and does not correct erroneous data association~\cite{montemerlo2002fastslam}.

    \subsubsection*{\textbf{Collaborative visual \ac{SLAM} (CoSLAM)}}
        
        This method~\cite{zou2013coslam} interacts in dynamic environments where live frames come from multiple cameras that can be independent and mounted on different points of view. As an overview, these cameras build a single global map, including the static background points and the foreground dynamic points. This set of points are the ones used to estimate the poses of all cameras, which should have overlapping fields of view. 
        CoSLAM is considered as one of the most efficient approaches which are able to get rid of false points caused by incorrect matching.
       
    \subsubsection*{\textbf{Large scale direct monocular \ac{SLAM} (LSD-SLAM)}} 
    
        It has been developed to allow the building of large-scale map environments~\cite{engel2014lsd}. Instead of using key points, this \ac{SLAM} uses the image intensities to track and build a map. 
        The method shows the advantage of allowing the mapping of large areas without extra computational power. 
    
    \subsubsection*{\textbf{Oriented FAST and rotated BRIEF \ac{SLAM} (ORB-SLAM)}} 
    
        It is a keyframe and \ac{ORB} feature based \ac{SLAM} algorithm~\cite{mur2015orb}. One of its greatest advantages is that it operates in real-time and large environments, being also able to close loops and re-localise from different viewpoints. Due to these significant contributions, it is the chosen algorithm for this project implementation. More details about the algorithm are described in section~\ref{sec:slam} 

\section{Object Recognition Framework}\label{sec:recognition}
% {\color{red} Should we change the section title to "Object recognition framework"? I do not like the proposed word here, because we are not really proposing it. Perhaps, we should also change "proposed" to something else in other sentences of this section.}
As previously explained, we want to allow the Pepper robot to be more autonomous and helpful in the household. One of our main objectives is to allow Pepper to recognise objects. Some of the classical object recognition algorithms show not to be efficient for some applications. For instance, Haar Cascades~\cite{viola2001rapid} requires a trade-off to be done between the efficiency in learning/training time and the output's accuracy.

Based on Lowe's paper~\cite{lowe1999object}, we introduce a robust \ac{SIFT}-based recognition algorithm for Pepper robot.
Our method is not only able to get rid of the long training time but also robust to rotation, scaling, perspective transformation among others. The robustness of this algorithm allows Pepper to recognise objects efficiently. 

\begin{figure}[h]
\centering
\includegraphics[width=0.8\linewidth]{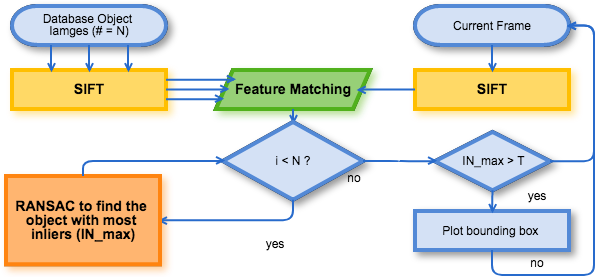}
\caption{Workflow of the object recognition algorithm.}
\label{fig:recog_workflow}
\end{figure}

The flow chart of the object recognition algorithm for Pepper is presented in Fig.~\ref{fig:recog_workflow}. The main steps of the method include feature extraction, feature matching, and decision making, which are described in the following sections.
% \begin{itemize}
% \item Feature extraction
% \item Feature matching
% \item Decision making
% \end{itemize}

\subsection{Feature extraction}\label{sec:featureExtraction}
    The very first step is to extract features from a given image database. This is also done every time a new frame arrives.
    Under the uniform recognition framework, we tested several feature extraction techniques including \ac{ORB}, \ac{SURF} and \ac{SIFT}.
    When applying \ac{ORB} and SURF the accurate detection of the output is acceptable. However, \ac{SIFT} offered the highest accuracy among them in all our testing rounds (see Section~\ref{sec:results}) so it was chosen as the primal feature extraction method.
    Both SURF and \ac{ORB} are kept as user options in the implementation and can be selected as the main feature descriptor.

\subsection{Feature matching}
    Once the \ac{SIFT} features of the images in the database and the current frame have been extracted, we need to know how many features are matched between the current frame and every image in the database.
    This will help the following decision-making step described in the next section. 
    
    We use the matching method proposed in~\cite{beis1997shape}, which is a kd-tree built for the rapid traversing of each feature in the current frame.
    From our intensive experiments, it has been shown that the kd-tree nearest neighbour matching algorithm significantly speeds up the recognition process (around three times faster than brute-force matching).
    Since the recognition run-time for each frame is within the range of the updating time ($200$ $ms$), the kd-tree matching enables Pepper to perform real-time object recognition along with the SLAM algorithm.
 
\subsection{Decision making}

    After the feature matching between the current frame and all the objects in the database, \ac{RANSAC}~\cite{fischler1981random} is applied to determine whether an object from the database exists in the current frame or not, and which object it is.
    
    We use \ac{RANSAC} to fit a homography transformation between the feature positions $P_C$ in the current frame and the object in the database $P_D$, and then calculate the number of inliers $N$. 
    If $N$ is larger than a threshold (empirically set to $15$), we make this object as a candidate.
    This process is repeated until the number of inliers of all database objects has been calculated.
    If the inlier numbers are all smaller than the threshold, we assume there is no object found in the current frame. Otherwise, the candidate with the most inliers is considered as the detected object. 
    Finally, if an object is detected, through the homography matrix acquired from \ac{RANSAC}, the bounding box of the best object position is drawn on the frame.

    More implementation details about the Pepper object recognition can be found in the \texttt{/pepper\char`_recog} folder of our GitHub repository.

\section{Simultaneous Localisation and Mapping}\label{sec:slam}
Nowadays, \ac{SLAM} is one of the most active research topics in Computer Vision and Robotics community. Many \ac{SLAM} algorithms have been developed as we have discussed in Section~\ref{sec:relatedwork}.
All of these algorithms share a common purpose but use different approaches depending on the available sensors. Some of them use lasers, LiDAR, cameras or RGB-D cameras (or a combination of different sensors). For instance, LSD and \ac{ORB} \ac{SLAM} algorithms are based on RGB(-D) camera and called visual \ac{SLAM}.

The required sensors to achieve the goal of this project have been described earlier. Among the tested SLAM algorithms, the \ac{ORB} \ac{SLAM} was better to cope with the limited sensor capabilities such as low frame rate, therefore being the implemented algorithm on Pepper. In this section, a brief introduction for \ac{ORB} features and ORB \ac{SLAM} is given as well as its extension \ac{ORB} \ac{SLAM}~2~\cite{murORB2} that uses an RGB-D camera.

% The integration of the algorithm with Pepper also will be discussed further.

\subsection{\ac{ORB} feature}\label{sec:orbFeature}
    Since we are working with the visual \ac{SLAM}, extracting features from the input video stream is commonly the essential step.
    Feature extraction is the base for the \ac{ORB} \ac{SLAM} algorithm that makes the robot understand the surrounding environment and localise itself, as well as closing the trajectory loop.
    The Oriented FAST and Rotated BRIEF feature~\cite{rublee2011orb}, known as the \ac{ORB}, is a state-of-the-art feature descriptor that is applied to our \ac{SLAM} algorithm.
    
    \ac{ORB} is built on the Features from Accelerated Segment Test (FAST) detector~\cite{rosten2006machine} and Binary Robust Independent
    Elementary Features (BRIEF) descriptor~\cite{calonder2010brief}.
    The original FAST detector provides neither the keypoint orientation nor the measure of the corners, which makes \ac{ORB} not rotation invariant. 
    Therefore, in the phase of keypoint detection of \ac{ORB}, the intensity centroid~\cite{rosin1999measuring} and the Harris corner measure~\cite{harris1988combined} are applied to remedy these disadvantages. 
    Similarly, although the BRIEF descriptor can be calculated efficiently and robust to additive illumination change, perspective distortion, etc., the performance of BRIEF diminishes significantly for the rotation over a few degrees. 
    To solve the weakness of BRIEF, the best BRIEF pairs with large variance and low correlation are learned from PASCAL VOC 2006 \cite{everingham2006pascal} and then the obtained BRIEF descriptors from the key points of the current image are steered based on the orientation of the key points. 
    
    \ac{ORB} is made up of the modified version of FAST and BRIEF that we mentioned before.
    It is rotational and scale invariant as well as robust to noise, and it has been shown that the performance of \ac{ORB} in many real-life applications is equivalent to or even slightly better than SIFT in some cases. 
    More importantly, \ac{ORB} is computationally inexpensive. 
    Compared with the costly SIFT, \ac{ORB} is at two orders of magnitude faster, which is suitable for our real-time \ac{SLAM} application.

\subsection{\ac{ORB} \ac{SLAM} -- Monocular}
    \ac{ORB} \ac{SLAM} mainly consists of three components running in parallel: \textit{tracking}, \textit{local mapping} and \textit{loop closing}.
    In the following sections, the main ideas of each component are described.
  
\subsubsection{Tracking}
    This process starts with the \textit{initialization} of the map. In the monocular case, the depth has to be computed using several images of the same scene by moving the camera in the horizontal/vertical direction with respect to the scene. The authors proposed a new method \cite{mur2015orb} for "structure from motion" estimation that combines two geometrical models for camera pose estimation: 
    \begin{itemize}
    \item Assumes the scene is planar and computes the corresponding homography matrix between two frames. 
    \item Assumes the scene is non-planar and computes the fundamental matrix.
    \end{itemize}
    
    Then the selection of the best model is computed using specific heuristics, and the camera pose will be estimated based on the selected model.
    Once the map has been initialised from several consecutive frames of a scene from different viewpoints, the \ac{ORB} features (key-points) are extracted from consecutive frames. 
    Note that the FAST corners are extracted at 8-scale levels, and the modified BRIEF descriptors are computed on the key points orientation.
    
    The camera pose is computed by searching the matches in a small area around each \ac{ORB} key point between the current frame and the previous one. The search is optimised by assuming that the camera motion has a constant velocity model. If there are not enough matches, the search is done on all map points near the points from the last observed frame. 
    In case the track is lost, the current key points are converted into bag-of-words features and traverse the predefined recognition bag-of-words database. This is applied to obtain the best matching keyframe. After that, the robot can be re-localised again.
    Moreover, an Efficient Perspective-n-Point (EPnP) algorithm \cite{lepetit2009epnp} along with RANSAC is applied to refine the estimation of the pose further.

\subsubsection{Local mapping}\label{sec:localmapping}
    The new keyframe is obtained as discussed in the last section, and to put the new map points, we need to find the positions of all the new points on the world coordinate.
    Instead of triangulating points only with the closest keyframes like PTAM, \ac{ORB}-\ac{SLAM} triangulates points with several neighbouring keyframes.
    As long as a pair of \ac{ORB} features have been matched, they can be triangulated.
    
    Sometimes wrong map points may appear. To ensure all the mapped points are the real ones, we should check if a map point remains in the map for a period of time. 
    The authors of \ac{ORB}-\ac{SLAM} use a method called \textit{pass culling test}, which means a key point can be put in the map only after the following two conditions are satisfied: make sure the key point can be found in at least $25\%$ of frames, and the key point should be seen in at least three keyframes.
    
    Finally, the local bundle adjustment will optimise the current keyframe. 
    The final pose optimisation is performed by the Levenburg-Marquart method.

\subsubsection{Loop closing}
    Loop closing is one of the most important contributions of the \ac{ORB}-\ac{SLAM} and also one of the reasons we chose it for Pepper's \ac{SLAM} task.
    Loop closing means when the robot is moving around the environment and then comes back to the starting point, the system should be able to connect the latest movement with the initial ones.
    In this case, the trajectory can be closed, and the map will be globally changed. 
    With the loop closing, the built map and the estimated robot trajectory
    are more accurate.
    
    The main idea of loop closing can be summarised in three steps: loop detection, similarity transformation computing and loop fusion. 
    First, a co-visibility consistency test is performed to check if a loop has been found. 
    Throughout the whole process of the \ac{SLAM}, we keep calculating the similarity between the current keyframe and all its neighbours in the co-visibility graph. The keyframe with the highest similarity score will be used to update the reference loop-closing frame.
    Second, if one keyframe satisfies the test in the first step, the RANSAC will be iteratively applied to calculate a similarity transformation containing: 3 translations, 3 rotations and 1 scaling parameter. When the candidate has enough number of inliers, we are sure the loop has been found.
    Third, with the similarity transformation matrix acquired from the last step, the map points in the current keyframe are reformed to the reference loop-closing keyframe. 
    The map points from all the neighbours of the current keyframe are also projected through the same transform. Therefore, all inliers from the last step are fused.
    
    The last step is to perform a global bundle adjustment. The only difference from section~\ref{sec:localmapping} is that optimising all the map points will be used for the bundle adjustment and refined.
    The illustrations of the loop closing can be found in Section~\ref{sec:results}.

\subsection{\ac{ORB} \ac{SLAM} -- RGB-D}
    As mentioned, the first step of \ac{ORB} \ac{SLAM} is the initialisation of the map, which requires several images of a scene from different viewpoints. However, this process takes a long time for Pepper robot, because with a rate of $5$ fps the sequence of images cannot provide a smooth parallax effect.
    
    In order to overcome this problem, an extension for Monocular \ac{ORB} \ac{SLAM} has been introduced in \cite{murORB2}, where the depth estimation has been replaced by the RGB-D camera. In this case, the initialisation process does not involve recovering the camera pose from several images. Instead, the first taken image by the camera can be directly used to initialise the map because the depth information for the key points is already there. Therefore, using an RGB-D camera speeds up significantly the initialisation process, which is very important when using a camera with a low frame rate as in the Pepper robot.

\section{Integration and Architecture} \label{sec:controlAndArch}
Our final objective was to combine the object recognition with \ac{SLAM}, i.e. while running \ac{SLAM} the robot can also identify the detected object's position and put a marker with the label on the map built by \ac{SLAM}.

In this section, we are going to show how we accomplished this task. Also, how the whole system is organised in order to make the robot, ROS and the two previously described algorithms work together. %The additional features integrated into the system will be introduced as well.

\subsection{System overview}
    Pepper robot comes with many built-in functions and its own operating system (OS) called NAOqi-OS. This is a Linux distribution based on Gentoo, and it is installed in Pepper's computer which is integrated on the robot. However, Pepper does not allow users to install third-party applications on its OS and requires to use its own Software Development Toolkit (SDK). In order to overcome this limitation, the Robotic Operating System (ROS) has been used in this project. ROS is a language and platform independent framework that gives users permission to create packages in a graph-based structure and provides a powerful tool for message sending/receiving between processes \cite{quigley2009ros}.

    \subsubsection*{\textbf{ROS -- NAOqi driver and plugin for Pepper}}
\begin{figure}[tb]
        \centering
        \includegraphics[width=1\linewidth]{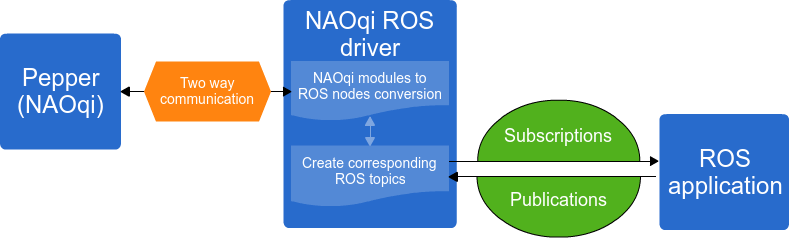}
        \caption{The diagram illustrating the way of communication of a ROS application through NAOqi ROS driver.}
        \label{fig:naoqiDriver}
    \end{figure}

    Despite the fact that the manufacturers of Pepper limit the access to the OS of Pepper, they provide a driver that can be used to link NAOqi and ROS together. This driver fetches all sensor data and creates ROS nodes and topics which publish the states of all the robot sensors. Moreover, the driver creates topics for controlling joints of the robot allowing other ROS applications to subscribe and publish standard ROS messages (e.g. Twist) to control the robot. The whole process of NAOqi-ROS communication is illustrated in Fig.~\ref{fig:naoqiDriver}. As can be seen in this figure, the main role of the NAOqi driver is converting NAOqi modules to ROS nodes.
    
    In addition to the NAOqi driver, there must be robot specific plugins that bring specific capabilities of the robot to ROS depending on the characteristics of the robot. For example, Pepper robot shares the same OS with other Aldebaran and Softbank robots, but each of these robots has different configurations such as the number of joints and types of sensors. In order to avail full robot capabilities, it is required to run a certain type of driver. To achieve this, \textit{pepper\_bringup} and \textit{pepper\_dcm\_bringup} plugins \cite{pepperBringup} have been used for the Pepper robot. The main difference between \textit{pepper\_bringup} and \textit{pepper\_dcm\_bringup} is that the former one does not block the autonomous life of the robot, whereas the latter turn that functionality off. In our experiments, we did not use the autonomous life, since it allows Pepper to imitate a human behaviour -- such as tracking human face, reacting to sudden loud noise, etc. -- which can bring inconvenience while running the algorithm.

\subsection{Implementation}

    Now we introduce how \ac{SLAM} and object recognition are combined using ROS. First of all, we have to mention that the \ac{ORB} \ac{SLAM} 2 algorithm that we used has been implemented in C++ programming language as a \textit{stand-alone} application, i.e. it can be used without ROS. For this reason, it does not use \textit{RViz} for showing the map, which is a default and convenient visualisation tool of ROS. Instead, it uses \textit{Pangolin}~\cite{pangolin}, which is a lightweight library for managing visualization and user interaction that wraps OpenGL~\cite{opengl} functions.
    
    \subsubsection{\textbf{Combining \ac{SLAM} with ROS}}
    To use \ac{ORB} \ac{SLAM} 2 in ROS, a ROS node was implemented to instantiate \ac{ORB} \ac{SLAM} 2 as an object. The created node subscribes to the topics where RGB and depth images are being published. Note that the \ac{ORB} \ac{SLAM} 2 with RGB-D expects an RGB-D camera, but Pepper has RGB and depth cameras separately. Accordingly, we made two separated subscribers for both modalities. We also have to make sure that the messages coming from these topics have the same \textit{timestamp}, because it is possible that some frames may be delayed or lost due to unexpected technical issues. Furthermore, we also make sure that the images from both cameras are correctly registered. The described architecture for \ac{SLAM} \& ROS is illustrated in Fig.~\ref{fig:architecture} (Block-A).
    
    % \subsubsection{Object recognition \& ROS}
    \subsubsection{\textbf{Combining object recognition with ROS}}
    In contrast to the \ac{ORB} \ac{SLAM} implementation, we implemented object recognition module as a ROS node, so the algorithm logic (Fig.~\ref{fig:recog_workflow}) is directly put inside the node. Then, we obtain images from Pepper's frontal camera and convert ROS raw image format to OpenCV image using CV-Bridge~\cite{cvbridge} package from ROS.
    
    The object position with respect to the camera coordinate (depth estimation) is computed in this node as well. The estimated depth, which will be used for marking objects on the map, is published as a topic. To publish the object name and its position in camera frame we created a custom ROS message that holds the following fields: 1) flag (boolean type) -- accepts \textit{true} when an object has been detected, \textit{false} otherwise; 2) depth (float type) -- estimated distance from the camera frame origin to the object; 3) name (string type) -- name of the object that has been detected. 

\subsubsection{\textbf{Marking objects on the map with homography}}
\begin{figure}[tb]
        \centering
        \includegraphics[width=0.9\linewidth]{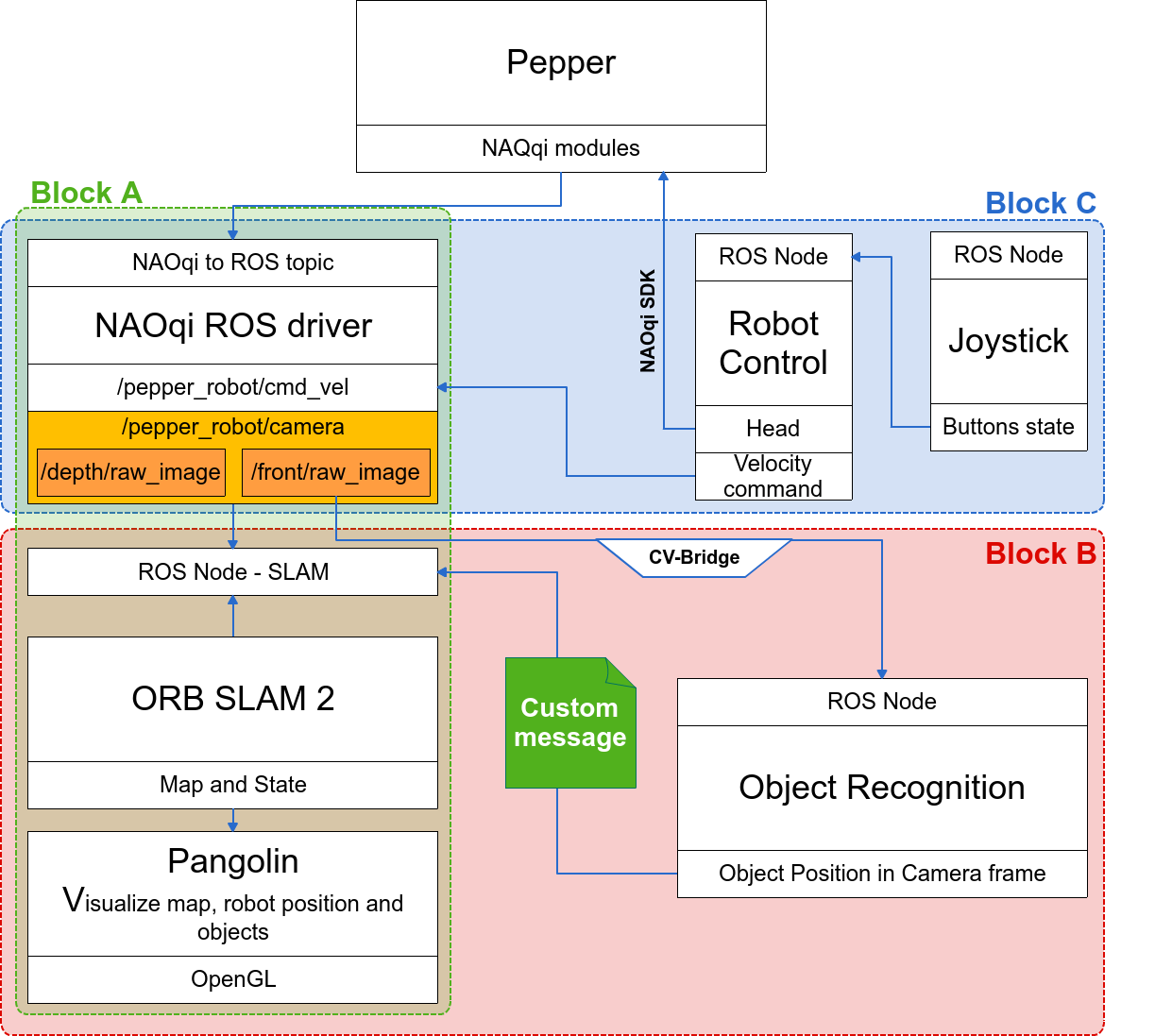}
        \caption{System architecture. Block A. Communication of NAOqi ROS driver with the ORB SLAM 2 module. Block B. Integration of ORB SLAM 2 with Object Recognition module. Block C. Robot control module and NAOqi ROS driver communication.}
        \label{fig:architecture}
    \end{figure}

    % \subsubsection{MARKING OBJECTS ON THE MAP}
    As we mentioned earlier when the object is detected the object recognition node publishes a custom message with a flag field set to \textit{true}. In order to put a marker with the name of the detected object, we created a subscriber to the custom message in the \ac{SLAM} node (Fig.~\ref{fig:architecture} (Block-B)). Since the map visualisation is independent of ROS, we cannot directly put markers on the map inside the \ac{SLAM} node.
    Therefore, we created a C++ class (we will refer to this class as \textit{Recognition.class} further) that represents the recognised objects in \ac{ORB} \ac{SLAM} 2 package. This class is also included in the ROS node.\
    When \ac{SLAM} node receives a message notifying that an object has been detected, we create an instance of the \textit{Recognition.class} with the parameters that came with the message.
    In order to process this kind of instances we modified the source code of \ac{ORB} \ac{SLAM} 2 to process \textit{Recognition.class} objects along with the RGB and depth images.
    % to process a set of custom messages for object recognition along with the RGB and depth images.
    % Mostly, the modification took place in the tracking process, which we have discussed in Section~\ref{sec:slam}. We also had to add functionality for Pangolin visualisation module to plot the positions and names of the objects. {\color{red}(SY: maybe give a bit more details of modifications here?) \color{blue}(KK: I would like to do it, but do not remember the details)}.
    
    Until now, we only know the positions of the objects w.r.t. The camera. Before plotting the object on the map, we have to find its position in the world frame. In order to do so, we obtained the pose (rotation + translation) of the camera when the object was being detected, which is described as the transformation matrix. Then, the position of the object is computed using the following equation:
    $$
        \underbrace{
            \begin{bmatrix}
                r_{1,1} & r_{1,2} & r_{1,3} & t_x\\
                r_{2,1} & r_{2,2} & r_{2,3} & t_y\\
                r_{3,1} & r_{3,2} & r_{3,3} & t_z\\
                0 & 0 & 0 & 1
            \end{bmatrix}
        }_{T_c}
        \times
        \underbrace{
            \begin{bmatrix}
                0\\ 0 \\ D_e \\ 1
            \end{bmatrix}
        }_{p_{object}}
        =
        \underbrace{
            \begin{bmatrix}
                r_{1,3}\times D_e + t_x\\
                r_{2,3}\times D_e + t_y\\
                r_{3,3}\times D_e + t_z\\
                1
            \end{bmatrix}
        }_{p_{world}}
    $$
    where $T_c$ - is the camera transformation matrix that shows how it is rotated and translated from the origin of the world frame; $p_{object}$ and $p_{world}$ - are object position in camera and world frames respectively; $D_e$ - is the estimated depth. Then we update the corresponding object coordinates with the new computed world frame coordinates.
    
    Now, by using the transformation matrix, the 3D position of the object with respect to the world coordinate has been found. As a result, we can directly put the marker with the name on that position in the map.
    
\subsection{Additional features}
    In this section, the additional features are shown that are essential for performing \ac{SLAM} and making the whole system faster and more practical. 

\subsubsection{\textbf{Robot control with a joystick}}
    The first thing that needs to be mentioned is the robot control. This is the main module that is used for moving the robot in an indoor environment for building the map. This task is executed with the help of a joystick. The usage of a joystick ensures full control of the robot for \ac{SLAM}. Additionally, by controlling the robot manually, we can assure that all the necessary areas of the environment are covered and put in the map.

    ROS provides a generic teleoperation tools \cite{teleopTools}, which is a simple library that reads commands from a joystick and publishes a vector with buttons state.
    In order to make it work with our robot, we created a controller ROS node, that subscribes to the joystick node. Then, depending on the pressed button, we define linear and angular velocities for the robot and send them to the /pepper\_robot/cmd\_vel topic.
    By sending velocity commands to that topic, we can control the robot base. However, it is worth mentioning this does not allow to control other joints of the robot.
    
    For controlling the robot head, we used NAOqi SDK inside our Robot Control node. First, we retrieve the current position of the head when a button, which was mapped to head movements, is pressed. Then, depending on the movement direction, we calculate the final position of the head (in degrees). Next, using the \textit{ALMotion} NAOqi module we send a command to the robot. Additionally, we programmed two more buttons that send the robot to \textit{Rest} and \textit{Active} status, which is implemented using NAOqi SDK as well.
    
    The general overview of the robot controlling component of the system is illustrated in Fig.~\ref{fig:architecture} (Block-C). The implementation details can be found in \texttt{/joy\char`_pepper/scripts/joypepper.py} in our GitHub repository.

    \subsubsection{\textbf{Map saving and loading}}
        Once the map of the environment has been built, it is important to be able to reuse it. Saving the map becomes an important task due to the short working period of Pepper's joints (e.g. overheat). The implementation of the \ac{ORB} \ac{SLAM} 2 does not provide a functionality that allows to save the built map and load an existing map. In order to fill this gap and allow Pepper to continue the map building process, we have included this feature to our system.
        
        First, a naive method has been implemented where we save all the key points, keyframes and corresponding bag of words for each keyframe of the map into a text file. To reuse it we load and parse this file. This method appears to be very slow and inefficient, due to the large file size. Moreover, the processes of writing/reading from a text file are known to be slow.
        
        Another way of solving this problem was saving all the instances of the C++ objects into a binary file, which is a well-known strategy in programming called \textit{serialization}. For \ac{ORB} \ac{SLAM} 2 there was already some research going on about this~\cite{serialization}, where serialisation and deserialization have been used for saving and loading the map. However, this has been implemented only for Monocular \ac{SLAM}, and we implemented it similarly for \ac{SLAM} with the RGB-D camera.
        More details can be found in the codes \texttt{Map.cc} \texttt{KeyFrame.cc} \texttt{MapPoint.cc} in our GitHub repository \texttt{/orb\char`_slam2/src}.

    \subsubsection{\textbf{Fast vocabulary loading}}
        For loop closing and camera relocalisation, the authors of \ac{ORB} \ac{SLAM} 2 used bag of words place recognition model~\cite{galvez2012bags}. This model uses a vocabulary of visual words, which have been built using a vast database of images. Whenever \ac{ORB} \ac{SLAM} 2 is launched it takes some time loading the vocabulary because the vocabulary is saved as a text file which contains more than a million lines. This issue makes the start-up process very slow, and therefore the serialisation for the vocabulary has been implemented similarly as in the map serialisation \cite{serialization}.
        The source code can be found \texttt{/tools/bin\char`_vocabulary.cc}.

    \subsubsection{\textbf{Object following and avoiding}}
        We also implemented an object following and avoiding functionality for Pepper. The main idea is to allow the robot to continuously track an object but also avoid it by keeping a certain distance when the object is too close. 
        
        The application works in the following manner: if the estimated depth distance from Pepper to the detected object is larger than $60$ $cm$, Pepper follows the object at a pre-defined constant speed. On the contrary, if the distance is smaller than $20$ $cm$ Pepper avoids it by going backwards. Moreover, we also want the detected object to be at the centre of the frame. In order to do so, we computed the displacement of the central point of the object from the frame centre. Then, depending on this displacement, we send an angular velocity command to the robot to minimise this difference.

\section{Results and discussion}\label{sec:results}
In this section we discuss the object recognition, SLAM and the integration. 
We will discuss what we have accomplished as well as the comparisons with other object recognition and SLAM methods. 
For the real demonstration, please check the link \url{https://youtu.be/evFsnWH_bpY}.

\subsection{Object recognition}

\begin{figure}[tb]
\centering
\subfigure[Runtime]{\label{fig:featurecomp1} \includegraphics[width=0.47\linewidth]{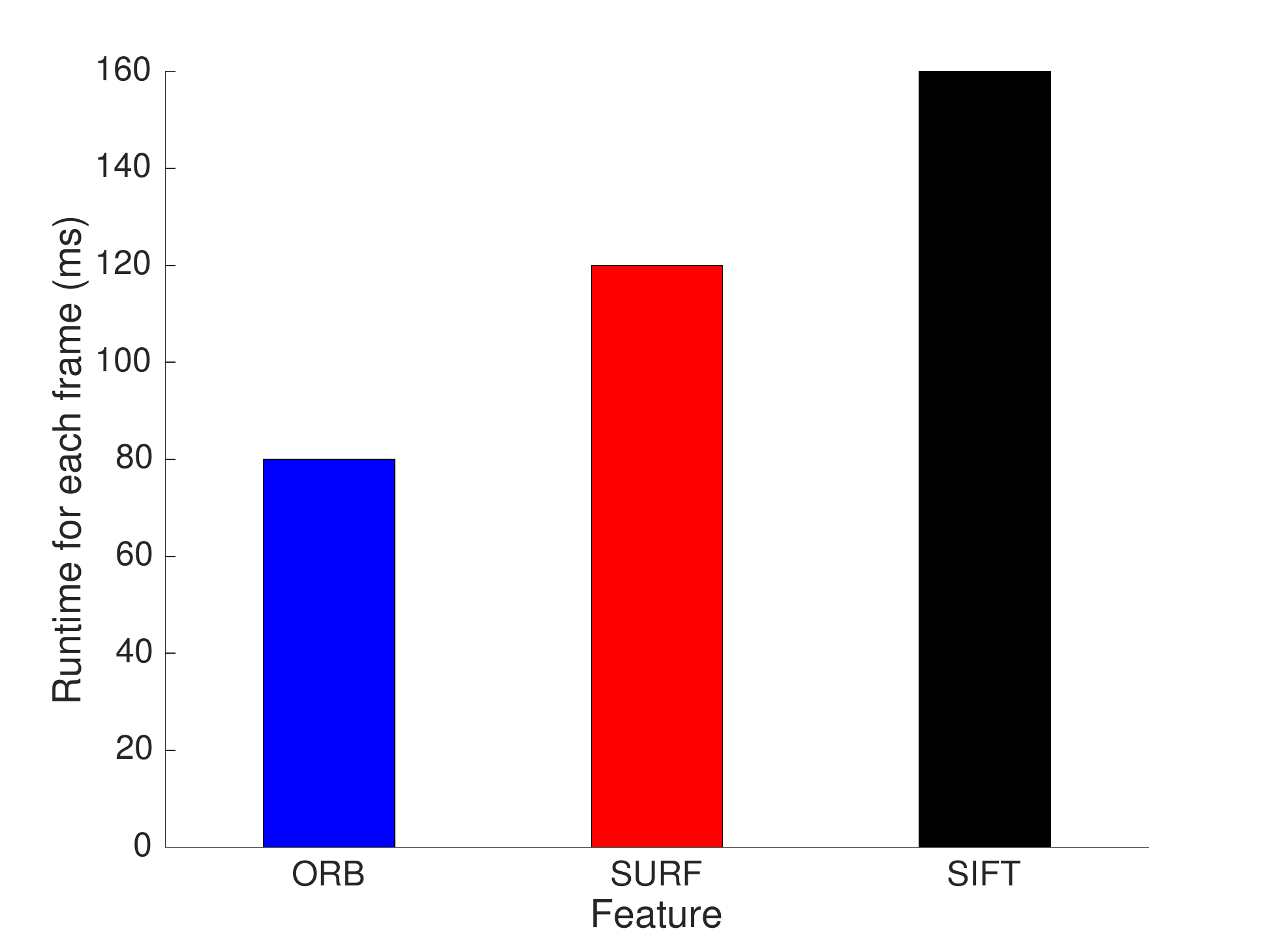}}
\subfigure[Accuracy rate]{\label{fig:featurecomp2} \includegraphics[width=0.47\linewidth]{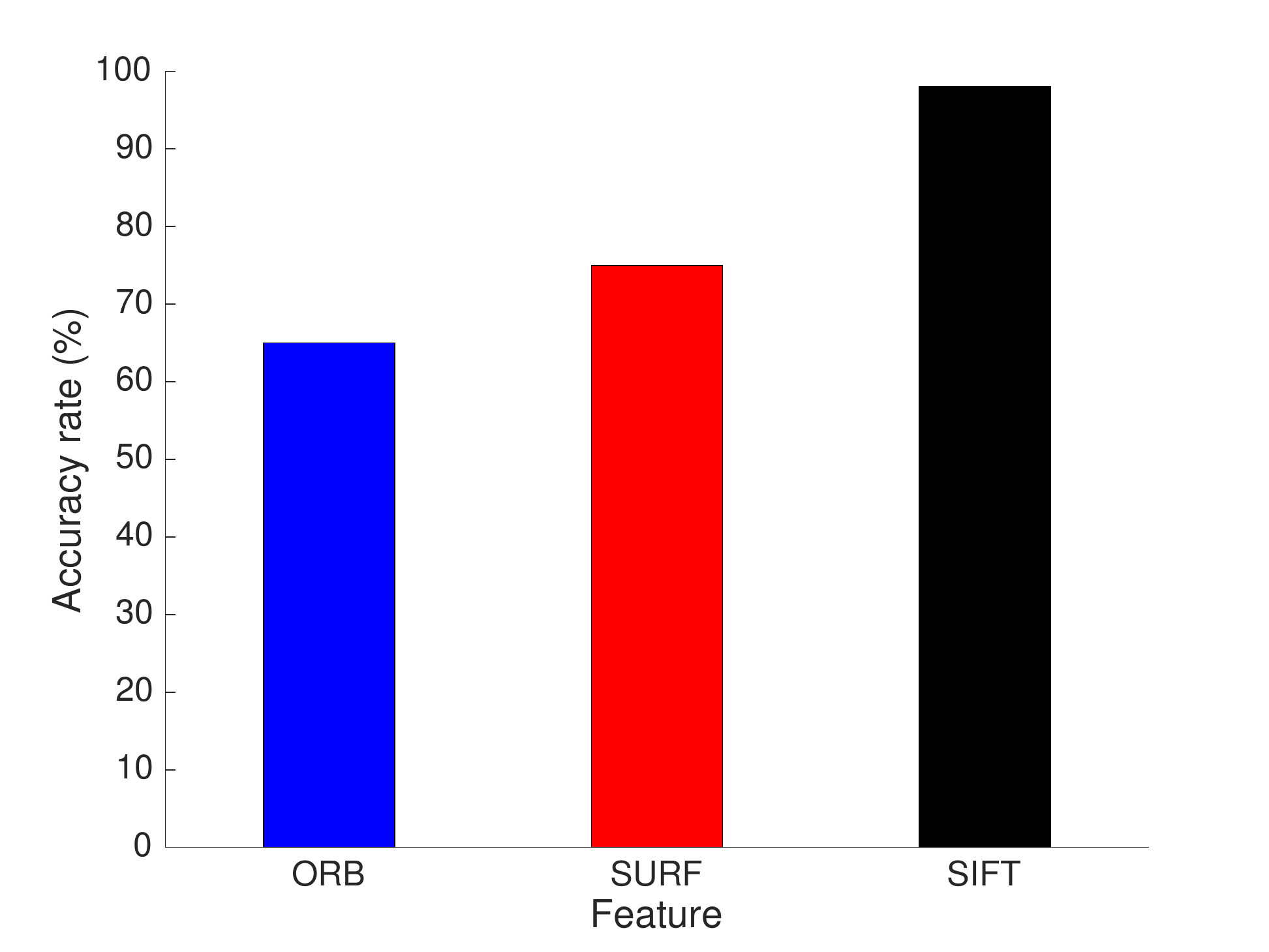}}
\caption{Comparison of the recognition performance with \ac{ORB}, SURF and SIFT. The implementation language is Python and 3 various objects (book, folder, T-shirt) are put in the database.}
\label{fig:featurecomp}
\end{figure}
% Because our approach for recognition is feature-based we will first compare the performance of the algorithm when applied with features.  
First of all, since we are using feature-based object recognition framework, the comparison of the recognition performance with various features should be discussed.
As illustrated in Fig.~\ref{fig:featurecomp}, we compare the performance of \ac{ORB}, SURF and SIFT under our recognition framework. In Fig.~\ref{fig:featurecomp1}, SIFT takes longer time than the other two but is under the acceptable updating time frame (200 ms).
Nonetheless, when comparing the recognition accuracy (Fig.~\ref{fig:featurecomp2}), the SIFT-based recognition achieves around $95\%$. Clearly, for the indoor usage, accuracy is more important than computational time, so SIFT is chosen as our primal feature.

%The accuracy definitely plays a more important role of our task for Pepper, hence, SIFT is chosen as our primal features extraction method for the object recognition task.

In Table 1, we compare our SIFT + NN + RANSAC method with the well-known Haar Cascade method~\cite{viola2001rapid}. 
As we can notice, our method outperforms the Haar Cascades almost in all the cases.
Haar Cascades method requires a long time to train one object, while our method does not require any training and needs only one image per object.
% not only does not need any training time, but also one single image for a certain object would be enough.
This makes the system more flexible and easy to use by allowing users to modify the database just by adding/removing images of objects.
% Therefore, instead of first learning how to train the data and to waste around 1 hour to train one object, users can take one picture of the desired image and then leave it in the database folder.
The performance of our method also appears to be much more consistent than the other method and barely has false alarms.

\begin{tcolorbox}[tabularx={X||c|c|c|c|c}, float=tb, boxrule=0.9pt, title= \centering Table 1: Comparison of Haar Cascade and our SIFT + NN + RANSAC]
 & Haar Cascades& Ours \\\hline\hline
Training time   &  {\color{red}\xmark}  & {\color{blue}\cmark}  \\\hline
Detection consistence    & {\color{red}\xmark}& {\color{blue}\cmark} \\\hline
False alarm    & {\color{red}\xmark}  & {\color{blue}\cmark} \\\hline
Rotation Invariant  & {\color{red}\xmark}& {\color{blue}\cmark}\\\hline
Detect with partial info &  {\color{red}\xmark}& {\color{blue}\cmark}\\\hline
A large number of objects & {\color{blue}\cmark} & {\color{red}\xmark}
\end{tcolorbox}

% Also, our method performs much more consistent than the other and barely has any false alarms.
% In comparison, the ed method has a more consistent performance than the Haar cascades, barely resulting in false alarms.

From our experiments, when recognising the same still object, we found out that the recognition accuracy of our method is almost 100\%. 
In contrast, the accuracy of Haar Cascades is less than 60\%, and false alarm and misdetection may happen even in between two consecutive frames. 
Indeed, for Haar cascades the more negative/positive samples we use for training, the better recognition rate we obtain.
However, the training time will also increase dramatically.

\begin{figure}[tb]
\centering
\subfigure{\label{fig:siftrotate1} \includegraphics[width=0.47\linewidth]{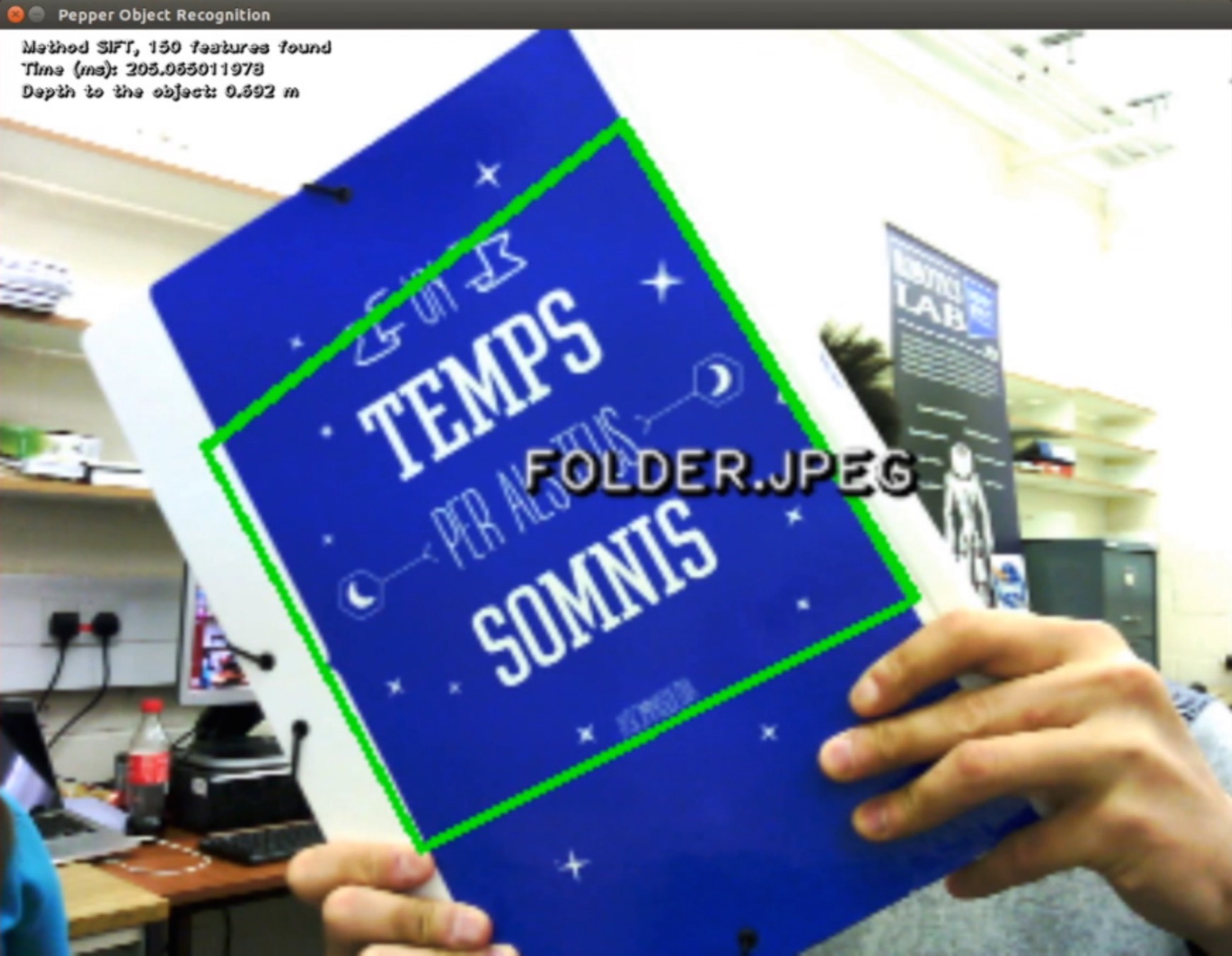}}
\subfigure{\label{fig:siftrotate2} \includegraphics[width=0.47\linewidth]{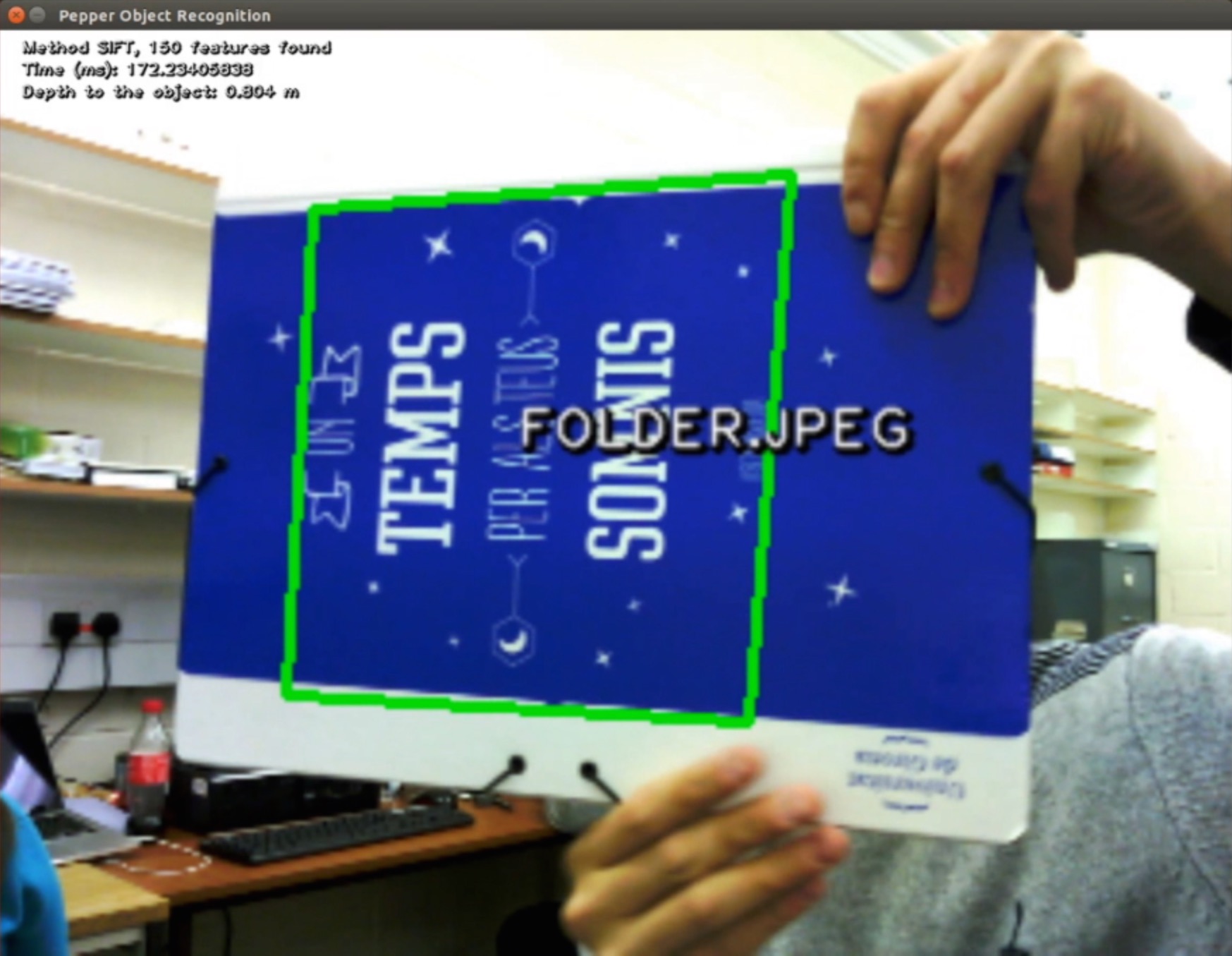}}
\caption{Illustrations for the rotation invariant of the object recognition.}
\label{fig:sift_rotate}
\end{figure}

Another important improvement of our method is enabling the object rotation-invariance. 
It turns out that the Haar-like feature does not evidently have the capacity of dealing with the rotated object unless a huge amount of samples with various angles have been used for training. Even though the training dataset is huge, we are not guaranteed with a decent result.
SIFT instead is mainly famed for the rotation-invariance property. 
Fig.~\ref{fig:sift_rotate} undoubtedly shows that our method can cope with all kinds of rotational movements. 

\begin{figure}[tb]
\centering
\subfigure{\label{fig:siftcover1} \includegraphics[width=0.41\linewidth]{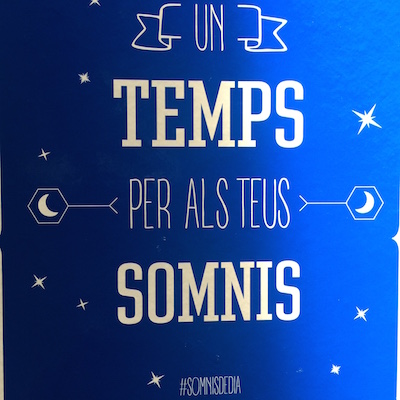}}
\subfigure{\label{fig:siftcover2} \includegraphics[width=0.52\linewidth]{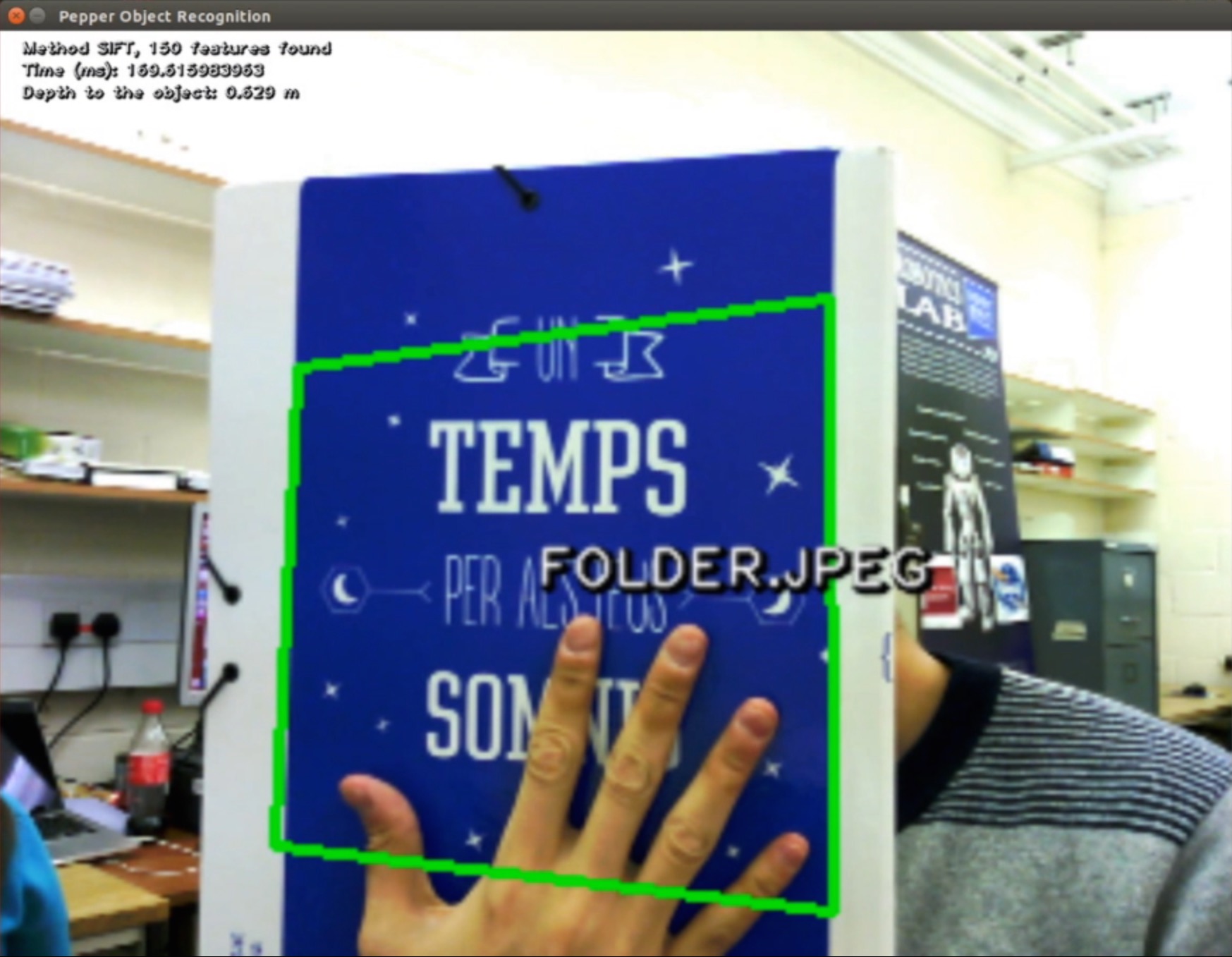}}
\caption{The object recognition can work with partial information.}
\label{fig:sift_obstruct}
\end{figure}

It is worth mentioning that our method can still recognise the objects properly with only partial details of an object, as shown in Fig~\ref{fig:sift_obstruct} with around $30\%$ of the folder covered.
% This improvement is due to the proper threshold that we set for the number of inliers. As long as the inliers number acquired from RANSAC is larger than 15, we say an object has been detected.
For Haar cascades, the object is not able to be recognised at all even if the covered portion is really small. 

%only a small part has been covered.

Finally, the main problem of our method is that the recognition will be slower when the number of the objects inside the database increases. 
% For future work, to improve this part, the Bag-of-Words + SVM training scheme can be included.
% In spite of this limitation we recognise that for a service robot which the main purpose is to perform a limited number of objects in the database indoors is enough.
However, with regard to a home service robot which always stays indoors, it is sufficient for Pepper to recognise only a limited number of objects.

\subsection{Visual \ac{SLAM} with object recognition}
The experiments have shown that the \ac{ORB} \ac{SLAM} 2 outperforms its predecessor and LSD-\ac{SLAM} in terms of initialisation and depth estimation accuracy. LSD-\ac{SLAM} provides the dense reconstruction of the scene, however, in an indoor environment, it will cause lots of noise due to the inaccurate depth estimation of the points. Therefore, \ac{ORB} \ac{SLAM} and \ac{ORB} \ac{SLAM} 2 algorithms showed better performance in mapping an indoor environment. It should also be mentioned that LSD-\ac{SLAM} is oriented for Large-Scale environments, whereas \ac{ORB} \ac{SLAM} can be applied for both outdoor and indoor environments.

\begin{tcolorbox}[tabularx={X||c|c|c|c}, float=tb, boxrule=0.9pt, title= \centering Table 2: Comparison of \ac{SLAM} algorithms]
 & LSD & ORB & ORB-2 & Ours\\\hline\hline
RGB-D support &  {\color{red}\xmark}    &  {\color{red}\xmark}    &  {\color{blue}\cmark}   &  {\color{blue}\cmark}  \\\hline
Fast initialization & {\color{red}\xmark}   & {\color{red}\xmark}   &  {\color{blue}\cmark}   &  {\color{blue}\cmark}  \\\hline
Accurate localization &{\color{red}\xmark}   & {\color{blue}\cmark}   &  {\color{blue}\cmark}   &  {\color{blue}\cmark}  \\\hline
Map saving \& reusing & {\color{red}\xmark}   & {\color{red}\xmark}   & {\color{red}\xmark}   & {\color{blue}\cmark}  \\\hline
Fast vocabulary load & {\color{red}\xmark}   & {\color{red}\xmark}   & {\color{red}\xmark}   & {\color{blue}\cmark} \\\hline
Recognition + SLAM & {\color{red}\xmark}   & {\color{red}\xmark}   & {\color{red}\xmark}   & {\color{blue}\cmark}
\end{tcolorbox}

Table 2 summarises the comparison of the \ac{SLAM} algorithms that we tried for the implementation as well as the improved \ac{SLAM} version that we are using to which we added extra features to the \ac{ORB} \ac{SLAM} 2 implementation.

It can be clearly seen that the final result we obtained is the best among the others. As it was explained before, the most important features that include map saving and reusing play a significant role while performing \ac{SLAM} with Pepper. Fast initialisation for the tracking process is also achieved by leveraging depth camera as well as the decrease of the launching time due to the serialisation of the vocabulary.

%%%%%% Loop Closing
\begin{figure}[tb]
\centering
\subfigure[Before]{\label{fig:loopclosing1} \includegraphics[width=0.4\linewidth]{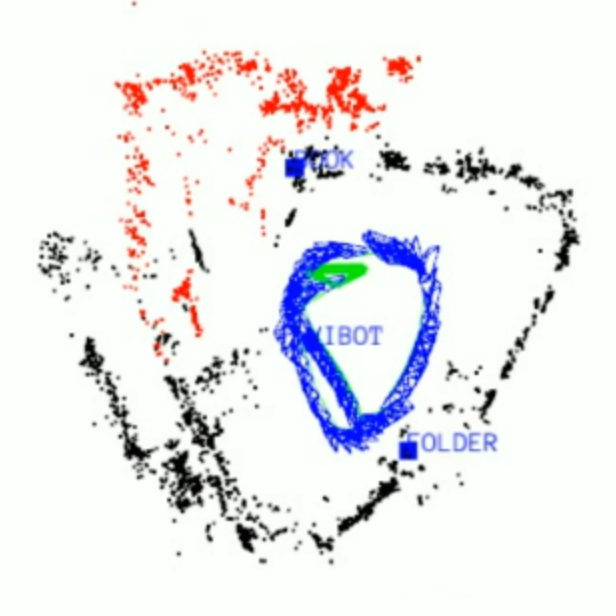}}
\subfigure[After]{\label{fig:loopclosing2} \includegraphics[width=0.4\linewidth]{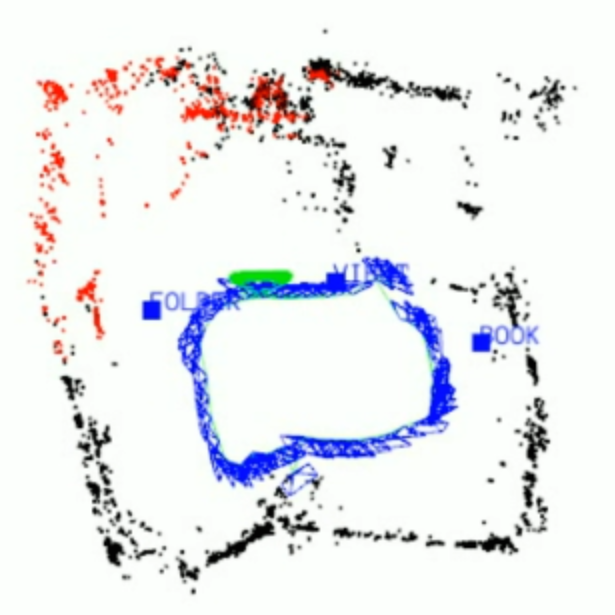}}
\caption{Illustration of the constructed map and localised objects (a) before and (b) after the loop closing.}
\label{fig:loopclosing}
\end{figure}
The results after running \ac{SLAM} + object recognition are illustrated in Fig.~\ref{fig:loopclosing}. Here we can observe that the map on the left (Fig.~\ref{fig:loopclosing1}) is a preliminary result that has been obtained before the loop closure. When the robot arrived at its initial position, the system closed the loop and reconstructed a map of the environment as well as the trajectory of the robot as shown in Fig.~\ref{fig:loopclosing2}. Blue markers on both images represent the inserted keyframes and the red (active), and black (inactive) points are the key points. 

\begin{figure}[tb]
\centering
\includegraphics[width=0.77\linewidth]{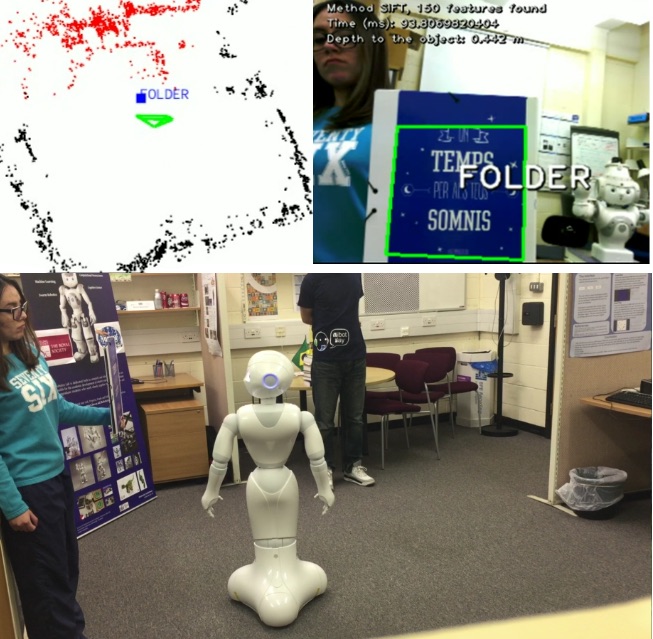}
\caption{Putting markers of detected object. Top-left: Map and the inserted marker; Top-right: Recognized object; Bottom: True location of the robot and object.}
\label{fig:localization}
\end{figure}

After the loop closure, we saved the map and reloaded it again to perform only localisation of the robot and to test object recognition and localisation on the map. The result of this test has been shown in Fig.~\ref{fig:localization}. From the top-left image, we can observe the previously built map and the robot position as well as the position and label of the recognised object, which is shown in the top-right image. The bottom image shows the robot and the part of the environment where we performed our tests.

\section{Final Remarks and Future Works}\label{sec:final-remarks}
% In this section, we summarise the results of our project and also introduce possible areas for future work.

One of the main aspects is that an innovative application integrating a robust object recognition algorithm with a modified \ac{ORB} \ac{SLAM} 2 was proposed. This system was implemented and successfully tested on the humanoid Pepper robot under the scheme of the European Robotics League.

As a summary, for the object recognition algorithm, SIFT features were extracted and then matched using kd-tree nearest neighbour search. 
Then, whether an object was recognised or not is decided through RANSAC. The algorithm has shown its robustness through its consistent detection, high accuracy without false alarm, and rotational invariance, etc.

Regarding the \ac{SLAM} application, we have modified and improved the open source \ac{ORB} \ac{SLAM} 2 in following ways: enabling the map saving and reusing it, accelerating the vocabulary loading and most importantly, integrating the object recognition.
The whole system is successfully working on Pepper despite the poor sensors, especially the low resolution and frame rate of the camera as well as the joint overheating problem.

%We successfully made the whole system works well on the Pepper even though Pepper's sensors are demonstrated to be very limited, especially the cameras resolution and frame rate and the joint overheating problem. 

Finally, some future works can easily be implemented on top of our proposed application, for example:
\begin{itemize}
\item Autonomous control of velocity while building the map.
\item Add path planning algorithms (e.g. Rapidly-exploring Random Tree, Rotational Plane Sweep, etc.).
\item Make Pepper go to the marked position of a certain object and be able to grasp it and take the object back to the initial position.
\end{itemize}

\section*{Acknowledgements}\label{sec:acknowledgement}

We want to thank our project supervisor Dr Mauro Dragone for his helpful guidance and support in the process, as well as Dr Yvan Petillot for his comments and feedback.

We also thank Raul Mur-Artal for his outstanding \ac{ORB} \ac{SLAM} 2 algorithm, Bence Magyar for his advice of using Joystick teleop, and Jos\'e Mar\'ia Sola Dur\'an for his object recognition code framework.

% Bibliography
\bibliographystyle{ieeetr}
\bibliography{Bibliography}

\begin{thebibliography}{10}

\bibitem{brown2013meet}
S.~Brown, ``Meet pepper, the emotion reading robot,'' {\em Health}, 2013.

\bibitem{erl}
eu{Robotics}, ``European robotics league.''
  \url{https://www.eu-robotics.net/robotics_league/}, 2019.

\bibitem{pepper243documentation}
ALdebaran, {\em Pepper Documentation {NAO}qi}.
\newblock ALdebaran Sofbank Group.

\bibitem{sung1998example}
K.-K. Sung and T.~Poggio, ``Example-based learning for view-based human face
  detection,'' {\em IEEE Transactions on pattern analysis and machine
  intelligence}, vol.~20, no.~1, pp.~39--51, 1998.

\bibitem{swain1991color}
M.~J. Swain and D.~H. Ballard, ``Color indexing,'' {\em International journal
  of computer vision}, vol.~7, no.~1, pp.~11--32, 1991.

\bibitem{schiele1996object}
B.~Schiele and J.~L. Crowley, ``Object recognition using multidimensional
  receptive field histograms,'' in {\em ECCV}, 1996.

\bibitem{koenderink1992generic}
J.~J. Koenderink and A.~J. van Doorn, ``Generic neighborhood operators,'' {\em
  IEEE Transactions on Pattern Analysis and Machine Intelligence}, vol.~14,
  no.~6, pp.~597--605, 1992.

\bibitem{linde2004object}
O.~Linde and T.~Lindeberg, ``Object recognition using composed receptive field
  histograms of higher dimensionality,'' in {\em ICPR}, 2004.

\bibitem{schneiderman2000statistical}
H.~Schneiderman and T.~Kanade, ``A statistical method for 3d object detection
  applied to faces and cars,'' in {\em CVPR}, 2000.

\bibitem{belongie2002shape}
S.~Belongie, J.~Malik, and J.~Puzicha, ``Shape matching and object recognition
  using shape contexts,'' {\em IEEE transactions on pattern analysis and
  machine intelligence}, vol.~24, no.~4, pp.~509--522, 2002.

\bibitem{viola2001rapid}
P.~Viola and M.~Jones, ``Rapid object detection using a boosted cascade of
  simple features,'' in {\em CVPR}, vol.~1, pp.~I--511, 2001.

\bibitem{lowe1999object}
D.~G. Lowe, ``Object recognition from local scale-invariant features,'' in {\em
  ICCV}, vol.~2, pp.~1150--1157, 1999.

\bibitem{Bay06surf:speeded}
H.~Bay, T.~Tuytelaars, and L.~V. Gool, ``{SURF}: Speeded up robust features,''
  in {\em ECCV}, pp.~404--417, 2006.

\bibitem{rublee2011orb}
E.~Rublee, V.~Rabaud, K.~Konolige, and G.~Bradski, ``{ORB}: An efficient
  alternative to sift or surf,'' in {\em ICCV}, 2011.

\bibitem{leutenegger2011brisk}
S.~Leutenegger, M.~Chli, and R.~Y. Siegwart, ``{BRISK}: Binary robust invariant
  scalable keypoints,'' in {\em ICCV}, 2011.

\bibitem{lecun1998gradient}
Y.~LeCun, L.~Bottou, Y.~Bengio, and P.~Haffner, ``Gradient-based learning
  applied to document recognition,'' {\em Proceedings of the IEEE}, vol.~86,
  no.~11, pp.~2278--2324, 1998.

\bibitem{girshick2015fast}
R.~Girshick, ``Fast {R-CNN},'' in {\em ICCV}, pp.~1440--1448, 2015.

\bibitem{liu2016ssd}
W.~Liu, D.~Anguelov, D.~Erhan, C.~Szegedy, S.~Reed, C.-Y. Fu, and A.~C. Berg,
  ``Ssd: Single shot multibox detector,'' in {\em ECCV}, 2016.

\bibitem{redmon2016you}
J.~Redmon, S.~Divvala, R.~Girshick, and A.~Farhadi, ``You only look once:
  Unified, real-time object detection,'' in {\em CVPR}, 2016.

\bibitem{montemerlo2002fastslam}
M.~Montemerlo, S.~Thrun, D.~Koller, B.~Wegbreit, {\em et~al.}, ``Fast{SLAM}: A
  factored solution to the simultaneous localization and mapping problem,'' in
  {\em AAAI}, pp.~593--598, 2002.

\bibitem{zou2013coslam}
D.~Zou and P.~Tan, ``Co{SLAM}: Collaborative visual {SLAM} in dynamic
  environments,'' {\em IEEE transactions on pattern analysis and machine
  intelligence}, vol.~35, no.~2, pp.~354--366, 2013.

\bibitem{engel2014lsd}
J.~Engel, T.~Sch{\"o}ps, and D.~Cremers, ``{LSD-SLAM}: Large-scale direct
  monocular {SLAM},'' in {\em ECCV}, pp.~834--849, 2014.

\bibitem{mur2015orb}
R.~Mur-Artal, J.~Montiel, and J.~D. Tard{\'o}s, ``{ORB SLAM}: a versatile and
  accurate monocular slam system,'' {\em IEEE Transactions on Robotics},
  vol.~31, no.~5, pp.~1147--1163, 2015.

\bibitem{beis1997shape}
J.~S. Beis and D.~G. Lowe, ``Shape indexing using approximate nearest-neighbour
  search in high-dimensional spaces,'' in {\em CVPR}, 1997.

\bibitem{fischler1981random}
M.~A. Fischler and R.~C. Bolles, ``Random sample consensus: a paradigm for
  model fitting with applications to image analysis and automated
  cartography,'' {\em Communications of the ACM}, 1981.

\bibitem{murORB2}
R.~Mur-Artal and J.~D. Tard\'os, ``{ORB-SLAM2}: an open-source {SLAM} system
  for monocular, stereo and {RGB-D} cameras,'' {\em IEEE Transactions on
  Robotics}, vol.~33, no.~5, pp.~1255--1262, 2017.

\bibitem{rosten2006machine}
E.~Rosten and T.~Drummond, ``Machine learning for high-speed corner
  detection,'' in {\em ECCV}, pp.~430--443, 2006.

\bibitem{calonder2010brief}
M.~Calonder, V.~Lepetit, C.~Strecha, and P.~Fua, ``Brief: Binary robust
  independent elementary features,'' in {\em ECCV}, 2010.

\bibitem{rosin1999measuring}
P.~L. Rosin, ``Measuring corner properties,'' {\em Computer Vision and Image
  Understanding}, vol.~73, no.~2, pp.~291--307, 1999.

\bibitem{harris1988combined}
C.~Harris and M.~Stephens, ``A combined corner and edge detector.,'' in {\em
  Alvey vision conference}, vol.~15, p.~50, 1988.

\bibitem{everingham2006pascal}
M.~Everingham, A.~Zisserman, C.~K. Williams, and L.~Van~Gool, ``The {PASCAL}
  visual object classes challenge 2006 (voc2006) results,'' 2006.

\bibitem{lepetit2009epnp}
V.~Lepetit, F.~Moreno-Noguer, and P.~Fua, ``{Epnp: An accurate O(n) solution to
  the PnP problem},'' {\em International journal of computer vision}, vol.~81,
  no.~2, pp.~155--166, 2009.

\bibitem{quigley2009ros}
M.~Quigley, K.~Conley, B.~Gerkey, J.~Faust, T.~Foote, J.~Leibs, R.~Wheeler, and
  A.~Y. Ng, ``{ROS}: an open-source robot operating system,'' in {\em ICRA
  workshop}, vol.~3, p.~5, 2009.

\bibitem{pepperBringup}
N.~Lyubova, ``Pepper bringup plugin.''
  \url{https://github.com/ros-naoqi/pepper_robot}, 2016.

\bibitem{pangolin}
S.~Lovegrove, ``Pangolin.'' \url{https://github.com/stevenlovegrove/Pangolin},
  2016.

\bibitem{opengl}
OpenGL, {\em OpenGL official documentation}.
\newblock OpenGL.

\bibitem{cvbridge}
P.~Mihelich and J.~Bowman, ``cv\_bridge.'' \url{http://wiki.ros.org/cv_bridge},
  2016.

\bibitem{teleopTools}
B.~Magyar, ``A set of generic teleoperation tools for any robot.''
  \url{https://github.com/ros-teleop/teleop_tools}, 2016.

\bibitem{serialization}
R.~Mur-Artal, ``{ORB SLAM 2} implementation, modified by a github user
  @poine.'' \url{https://github.com/poine/ORB_SLAM2}, 2016.

\bibitem{galvez2012bags}
D.~G{\'a}lvez-L{\'o}pez and J.~D. Tard{\'o}s, ``Bags of binary words for fast
  place recognition in image sequences,'' {\em IEEE Transactions on Robotics},
  vol.~28, no.~5, pp.~1188--1197, 2012.

\end{thebibliography}
\end{document}